\documentclass[twoside]{article}
\usepackage[accepted]{aistats2018}
\usepackage{cuted}

\usepackage[sort&compress,square,comma,authoryear]{natbib}
\usepackage[utf8]{inputenc} % allow utf-8 input
\usepackage[T1]{fontenc}    % use 8-bit T1 fonts
\usepackage[draft]{hyperref}       % hyperlinks
\usepackage{url}            % simple URL typesetting
\usepackage{booktabs}       % professional-quality tables
\usepackage{amsfonts}       % blackboard math symbols
\usepackage{nicefrac}       % compact symbols for 1/2, etc.
\usepackage{microtype}      % microtypography

\usepackage{algorithm}% http://ctan.org/pkg/algorithm
\usepackage{algpseudocode}% http://ctan.org/pkg/algorithmicx

\usepackage{caption}
\usepackage{subfig}

\newlength\myindent
\usepackage{mlmacros}
\usepackage{amsthm}
\usepackage{amsmath} % fleqn for left alignment of math blocks (as in classicthesis)
\usepackage{amssymb}
\usepackage{cancel}
\usepackage{cleveref}

\probdists{p,q}
\probdists[policy,source,transition,plan,initial]{\pi,\omega,p,q,\rho}
\probdists[optimalpolicy,optimalsource]{\pi^{\star},\omega^{\star}}
\probdists[recog,recogmeas,recogtrans]{q,q_{\text{meas}},q_{\text{trans}}}
\MkProbDist{empirical}{\hat{p}}
\variables[obs,innov,state,action]{x,e,z,u}

\newcommand{\sourceplanningpars}{\theta}
\newcommand{\sourcepars}{\phi}
\newcommand{\planningpars}{\psi}
\newcommand{\policypars}{\chi}
\newcommand{\dataset}{\mathcal{D}}
\newcommand{\modelpars}{\lambda}

\newcommand{\succstate}{{\bstate^\prime\!}}
\newcommand{\nowstate}{\bstate}
\newcommand{\nowaction}{\baction}

\newcommand{\margtrans}{\p{\succstate}{\nowstate}}
\newcommand{\jointtrans}{\p{\succstate, \nowaction}{\nowstate}}
\newcommand{\sourcetrans}{\source{\nowaction}{\nowstate}}
\newcommand{\planning}{\p{\nowaction}{\succstate, \nowstate}}
\newcommand{\apxplanning}{\q{\nowaction}{\succstate, \nowstate}}

\newcommand{\transmean}{\mu_{\text{trans}}}
\newcommand{\priortransstd}{\sigma_{\text{trans}}}
\newcommand{\priortransvar}{\sigma^2_{\text{trans}}}

\newcommand{\inftransstd}{\bar{\sigma}_{\text{trans}}}
\newcommand{\inftransvar}{\inftransstd^2}

\newcommand{\measuremean}{\mu_{\text{meas}}}
\newcommand{\measurestd}{\sigma_{\text{meas}}}
\newcommand{\measurevar}{\measurestd^2}

\newcommand{\recogmean}{\mu_q}
\newcommand{\recogstd}{\sigma_q}
\newcommand{\recogvar}{\recogstd^2}

\newcommand{\emp}{\mathcal{E}}
\newcommand{\empbound}{\mathcal{\hat{E}}}

\newcommand{\pd}[2]{\frac{\partial{#1}}{\partial{#2}}}
\newcommand{\Cost}{\mathcal{C}}
\newcommand{\modelloss}{\mathcal{L}_{\text{MODEL}}}     

\newcommand{\elbo}{\mathcal{L}_{\text{ELBO}}}

\DeclareDocumentCommand\mi{oG{}G{}}{\ensuremath{\operatorname{\mathcal{I}}\IfNoValueOrEmptyTF{#1}{}{_{#1}}\IfNoValueOrEmptyTF{#2}{}{\mleft(#2 \IfNoValueOrEmptyTF{#3}{}{\mid #3}\mright)}}}

\DeclareDocumentCommand\mibound{oG{}G{}}{\ensuremath{\operatorname{\mathcal{\hat{I}}}\IfNoValueOrEmptyTF{#1}{}{_{#1}}\IfNoValueOrEmptyTF{#2}{}{\mleft(#2 \IfNoValueOrEmptyTF{#3}{}{\mid #3}\mright)}}}

\usepackage{tikz}

\begin{document}


\runningauthor{Karl, Soelch, Becker-Ehmck, Benbouzid, van der Smagt, Bayer}

\twocolumn[

\aistatstitle{Unsupervised Real-Time Control through Variational Empowerment}

\aistatsauthor{Maximilian Karl \And Maximilian Soelch \And Philip Becker-Ehmck}
\aistatsauthor{Djalel Benbouzid \And Patrick van der Smagt \And Justin Bayer}
\aistatsaddress{ \{first name dot last name\}@volkswagen.de \\ AI Research, Data:Lab, Volkswagen Group, Munich, Germany}

]

\begin{abstract}
  We introduce a methodology for efficiently computing a lower bound to \textsl{empowerment},
allowing it to be used as an unsupervised cost function for 
policy learning 
in real-time control.
Empowerment, being the channel capacity between actions and states,
maximises the influence of an agent on its near future.
It has been shown to be a good model of biological behaviour in the absence of an extrinsic goal.
But empowerment is also prohibitively hard to compute, 
especially in nonlinear continuous spaces.
We introduce an efficient, amortised method for learning empowerment-maximising policies.
We demonstrate that our algorithm can reliably handle continuous dynamical systems using system dynamics learned from raw data.
The resulting policies consistently drive the agents into states where they can use their full potential.
\end{abstract}

%!TEX root = ../nips_2017.tex
\section{Introduction}
Sequential decision-making is one of the key problems in machine learning:
an agent interacting with an environment tries to optimise its behaviour based on accumulated cost or reward over time.
The framework is powerful due to its generality. 
A wide range of problems can be formalised, such as 
robot control \citep{kober2012reinforcement},
stock trading \citep{tsitsiklis1999optimal,tsitsiklis2001regression,li2009learning}, 
targeted marketing \citep{abe2004cross}, 
learning to play games \citep{tesauro1994td,silver2007reinforcement,silver2016mastering}, 
predictive maintainance \citep{gosavi2004reinforcement}, 
process automation \citep{zhang1995reinforcement} 
or electrical power systems \citep{ernst2005approximate}.

Despite recent successes in finding policies for increasingly hard tasks, substantial engineering effort is invested into the proper design of cost functions.
Finding a cost function that encourages a robot to walk is all but trivial.
Typically, we fall back to rewarding the consequences of the desired behaviour, such as moving fast.
And even if a cost function might appear obvious in some cases (e.g.\ profit in the case of stock trading), it often has to be regularised to be, e.g., risk-averse. 
This leads to an alternating process of reviewing the behaviour of an agent and adapting the cost function accordingly to achieve the desired behaviour.

There are various reasons for this.
For one, the problem is often ill-posed in the sense that it has many solutions: unconstrained degrees of freedom lead to behaviour that does not reflect the user's intent, even when a cost is minimised. 
Further, some cost functions---such as using the Euclidean distance to a goal state that a robot arm is supposed to reach---do not reflect the dynamics of the system.

In this work we advocate the use of unsupervised control techniques: each state's value is based solely on the system dynamics, without the notion of an agent or an externally supplied reward function. 
In particular, we opt for \emph{empowerment} \citep{klyubin2005empowerment,salge2014empowerment}, an information-theoretic formulation of the agent's influence on the near future. 
The value of a state, its empowerment value, is given by the maximum mutual information between a control input and the successor state the agent could achieve.

For instance, when balancing an inverted pole, states with high empowerment are those where the pole is standing upright.
For a biped, standing upright leads to highest empowerment.
States of low empowerment are those where the action has no effect on the state, and should be avoided. Links to biological behaviour have been shown \citep{klyubin2005empowerment,klyubin2008options}.

Yet empowerment is known to be notoriously hard to calculate.
Among other reasons, it requires integrating over all possible actions at all states, which is particularly difficult in continuous systems.
Most known methods circumvent these issues by operating in discrete action state spaces and only few works manage to scale to high-dimensional continuous state spaces \citep{mohamed2015variational,gregor2016variational,karl_efficient_2015}.

The contribution of this paper is an efficient, amortised method to estimate a lower bound to the empowerment of states in arbitrary continuous high-dimensional Markovian systems via variational methods.
In practice, we also illustrate the behaviour of policies maximising empowerment in non-Markovian systems. 
We experimentally verify the method on a range of tasks, leveraging a recent method for unsupervised learning of system dynamics from data via Deep Variational Bayes Filters (DVBF) \citep{karls2016deep}.
This allows for effective model-based learning of policies driven by empowerment,
using unsupervised approaches only.

%!TEX root = ../effemp.tex
\section{Methods}
\label{sec:methods}
We consider a dynamical system over sequences $\bstateTs$ subject to control inputs $\bactionTs$.
The first state $\bstate_1$ is drawn from the \emph{initial state distribution} $\initial$; subsequently control inputs and successor states are drawn alternatingly from some \emph{policy} $\policy{\baction_t}{\bstate_t}$  and the \emph{transition distribution} $\p{\bstate_{t+1}}{\bstatet, \bactiont}$.
The complete likelihood is hence given as
\eq{
    \p{\bstateTs, \bactionTs} = \initial{\bstate_1} \prod_{t=1}^T\p{\bstate_{t+1}}{\bstate_t, \baction_t}\,\policy{\baction_t}{\bstate_t}. \numberthis \label{eq:joint_dynamics}
}
Note that we consider the state transition to be Markovian, \ie $\bstate_{t+1}$ is conditionally independent of $\bstate_{1:t-1}$ and $\baction_{1:t-1}$ given $\bstate_t$ and $\baction_t$. In the following we also use $\nowstate$ and $\succstate$ to denote $\bstate_t$ and its successor state $\bstate_{t+1}$ respectively.

\subsection{A bound on empowerment}
\label{sec:methods:efficient_estimation_of_empowerment}
The empowerment value of a state $\nowstate$ has been defined as the channel capacity between the action $\baction$ and the following state $\succstate$  \citep{klyubin2005empowerment},
\eq{
    \emp(\nowstate) = \max_{\source}\, \mi{\succstate, \baction}{\nowstate}. \numberthis \label{eq:empowerment}
}
Here, $\mathcal{I}$ is the mutual information. 
In this context $\source$ is a policy, a \emph{source distribution}.
It is distinct from the empowerment-maximising policy~$\policy$.
By definition, empowerment is the maximum information (measured in \emph{nats}) an agent can emit to its environment by changing the state through its actions.
We will refer to the maximising policy $\optimalsource = \arg\max_{\source}\, \mi{\succstate, \baction}{\nowstate}$ as the \emph{optimal source}.
Note that this problem can be interpreted as a stochastic optimal-control problem \citep{stengel2012optimal}, where we are trying to find a control policy that maximises an information-theoretic quantity of the environment.

To tackle empowerment, consider the mutual information in \cref{eq:empowerment}.
It is defined as 
\eq{
    \mi{\succstate, \nowaction}{\nowstate} 
        :=& \kl{\jointtrans}{\margtrans\,\source{\nowaction}{\nowstate}} \\
        =& \iint\!  \jointtrans \ln \frac{\jointtrans}{\margtrans \,\sourcetrans}\dint{\succstate} \dint{\nowaction},
}
where we use a source distribution $\source$ as policy. 
The intractability of \cref{eq:empowerment} is due to the \emph{marginal transition} 
\begin{align*}
	\margtrans = \int \sourcetrans \, \jointtrans \dint{\nowaction}.
\end{align*}
We avoid the integral over all actions by switching to the \emph{planning distribution} $\planning$:
\eq{
    \mi{\succstate, \nowaction}{\nowstate} \!
        &= \!\!\iint \! \jointtrans\! \ln \frac{\cancel{\margtrans}\, \planning}{\cancel{\margtrans}\,\sourcetrans}\dint{\succstate} \dint{\nowaction}.
}
While the planning distribution is equally intractable, it is easier to employ a variational approximation $\apxplanning$. This leads to the bound introduced by \cite{barber2003im}:
\eq{
    \mi{\succstate, \nowaction}{\nowstate} 
        &\ge  \iint  \jointtrans \ln \frac{\apxplanning}{\sourcetrans}\dint{\succstate} \dint{\nowaction} \numberthis \label{eq:vi_bound}\\
        &=: \mibound{\succstate, \nowaction}{\nowstate}.
}
The gap in \cref{eq:vi_bound} can be expressed as
\eq{
	\mi - \mibound 	=   \expc[\succstate \sim \margtrans]{\kl{\planning}{\apxplanning}},\numberthis\label{eq:gap}
}
cf.\ \cref{app:bound} for details. 
Hence, as $\mi$ is constant \wrt $q$, maximising the bound $\mibound$ \wrt $q$ and $\source$ simultaneously maximises the mutual information while keeping the variational approximation $\apxplanning$ close to the true planning distribution $\planning$.

If the variational approximation is learnt well and chosen to be sufficiently expressive to represent the true planning distribution, the bound will be tight.
Quite a few means to improve upon variational approximations have been proposed recently, e.g.\ by \cite{rezende2015flows,burda2015importance,kingma2016improving}.

Replacing $\mi$ with $\mibound$, we achieve a lower bound on empowerment:
\eq{
    \empbound(\nowstate) :=& \max_{\source}\mibound{\succstate, \nowaction}{\nowstate} \numberthis \label{eq:emp_bound} \le \emp(\nowstate).
}

\subsection{Efficient mutual information optimisation}
\label{sub:efficient_mi}
\Cref{eq:emp_bound} establishes a lower bound on empowerment which relies on the estimator $\mibound$ from \cref{eq:vi_bound} for mutual information.
This section provides a new, efficient method to exploit this bound.

The key assumption is that we can efficiently compute
\eq{
	\pd{}{\sourceplanningpars}\mibound{\succstate, \nowaction}{\nowstate} = 	\pd{}{\theta}\expc[\jointtrans]{ \ln \frac{\apxplanning}{\sourcetrans} },
}
where $\sourceplanningpars$ denotes the joint of parameters $\source$ and $q$.

If both the system dynamics and the source can be sampled efficiently, we can estimate the gradients of the mutual information bound via Monte-Carlo sampling and the reparametrisation trick \citep{kingma2014auto,RezendeMW14}, \ie 
\eq{
	\pd{}{\sourceplanningpars} \mibound{\succstate, \nowaction}{\nowstate}
	\approx \frac{1}{N} \sum_{n=1}^{N} \pd{}{\sourceplanningpars}\left[ \ln \q{\nowaction^{(n)}}{\nowstate, \succstate\,^{(n)}} \right.\\ 
    \left. -\ln \source{\nowaction^{(n)}}{\nowstate} \right],
}
where $\nowaction^{(n)} \sim \source{\nowaction}{\nowstate}$ and $\succstate\,^{(n)} \sim \p{\succstate}{\nowstate, \nowaction^{(n)}}$ as in \cref{eq:joint_dynamics}.
Here, we assume that $\pd{\succstate}{\nowaction}$, $\pd{\succstate}{\nowstate}$ and $\pd{\nowaction}{\sourceplanningpars}$ can be efficiently estimated.

\cite{mohamed2015variational} report a model-free optimisation of the same bound $\mibound$.
In this case optimising $\mibound$ for $\source$ collapses to maximising the entropy of $\source$, which requires them to constrain said entropy.
In contrast, our sampling process couples $q$ and $\source$:
the samples used to evaluate $q$ are drawn using $\source$ and system dynamics---in particular no rollouts---and the entropy of $\source$ is balanced by the negative entropy of $\q$. \cite{mohamed2015variational} create this coupling by minimising a squared error containing $\q$ and the approximation of $\source$.

Our methodology allows for principled joint optimisation of planning and source distribution via backpropagation \citep{rumelhart1986learning}, for which, to the best of our knowledge, no experimental results have been reported previously.

\subsection{Exploiting empowerment}
\label{sec:methods:empowerment_maximising_agents}
The findings of the previous section allow for efficient learning of the optimal source $\optimalsource$, which in turn allows us to estimate empowerment.
The main goal of this is not to obtain a certain policy, but to estimate the empowerment value of states.
As the empowerment value can be perceived as an unsupervised value of each state, it can be used to train agents which proceed towards empowering states.
We hence use negative empowerment as a cost function in a stochastic optimal-control framework \citep{stengel2012optimal}.
Let such an agent be represented by a policy $\policy$ with parameters $\policypars$.
We then train it to minimise the cost function
\eq{
     \Cost(\policy) := -\expc[\policy]{\sum_{t=1}^T \empbound(\bstate_t)}  
    \approx& \, -\frac{1}{M} \sum_{m=1}^M \sum_{t=1}^T \empbound\left(\bstatet^{(m)}\right) \numberthis \label{eq:empowering_cost}.
}
The right-hand side approximation of $\Cost(\policy)$ is computed by a two-stage sampling process:
the samples $\bstatet^{(m)}$ are sampled according to the system dynamics model and the policy $\pi$.
Within the evaluation of $\empbound\bigl(\bstatet^{(m)}\bigr)$ according to \cref{sub:efficient_mi}, we take $\bstatet^{(m)}$ as a seed for further sampling one step according to the source distribution $\source$.
For protecting the policy from getting stuck at suboptimal solutions we introduce a KL divergence between the policy $\policy$ and a standard normal distribution.
A parameter $\beta$ weights the cost against this regulariser.
The procedure is summarised in \cref{alg:training}.
The assumptions of \cref{sub:efficient_mi} allow us to efficiently estimate all gradients of $\cost{\policy}$ via, e.g.,
reparametrisation.

We want to stress two properties where our method differs.
Firstly, the availability of a differentiable model lets us propagate through the transitions, enabling the agent to trade short term cost for future reward.
Secondly, the time horizon $T$ is a hyper-parameter only of the training of $\policy$, and setting it to a different value does not require re-estimation of $\apxplanning$ and $\empbound$.

This allows us to amortise the exploitation of empowerment over an arbitrary horizon into a single, fixed-computation time policy $\policy$, granting real-time applicability.

\begin{algorithm*}[t!]
		\caption{Joint training of policy $\policy$, source $\source$ and approx.\ planning distribution $\apxplanning$}
	\begin{algorithmic}
		\Require dynamics $\p{\bstate_{t+1}}{\bstate_t, \baction_t}$, cumulation horizon $T$, initialisations for $\policy, \source,$ and $\q$
		\Repeat 
			\For{$m=1:M$}\Comment{Estimate cumulated empowerment $\Cost$, \cref{eq:empowering_cost}}
				\State {$\policy$-step: Sample dynamics with policy ${\policy}$:}
				\State \hspace{\algorithmicindent}{$ \bstate_{1:T}^{(m)}, \baction_{1:T-1}^{(m)} \sim \transition{\bstate_1} \prod_{t=1}^{T-1} \transition{\bstate_{t+1}}{\bactiont,\bstatet} \policy{\bactiont}{\bstatet}$ }
				\For{$t=1:T$}
					%\State {estimate $\empbound\left(\bstatet^{(m)}\right)$:}
					\State $\source$-step: Sample one step with policy ${\source}$: 
					\State \hspace{\algorithmicindent}{$\bactiont^{\source} \sim \source{\bactiont}{\bstatet^{(m)}}, \bstate_{t+1}^\omega \sim \transition{\bstate_{t+1}}{\bactiont^\omega,\bstatet^{(m)}}$} 
					\State {$\hat{\mathcal{I}}_t^m \leftarrow \ln {\plan{\bactiont^\omega}{\bstate_{t+1}^\omega, \bstatet^{(m)}}{\theta}} - \ln\source{\bactiont^\omega}{\bstatet^{(m)}}$
					\Comment{One-shot MI, \cref{eq:vi_bound,eq:emp_bound}}}
				\EndFor
			\EndFor
			\State {$\Cost \leftarrow \sum_m\sum_t \left[ \beta \hat{\mathcal{I}}_t^m + \kl{\policy{\bactiont^{(m)}}{\bstatet^{(m)}}}{\policy_0} \right]$\Comment{Cost and KL towards policy prior $\policy_0$}}
			\State {$\theta \leftarrow \theta + \lambda_\theta\nabla_{\!\theta}\, \mathcal{C}$ \Comment{gradient ascent in $\source$ and $q$ with some learning rate $\lambda$}}
			\State {$\chi \leftarrow \chi + \lambda_\chi\nabla_{\!\chi} \,\mathcal{C}$ \Comment{gradient ascent in \policy\ with some learning rate $\lambda$}	}	
		\Until{convergence}

	\end{algorithmic}

	\label{alg:training}
\end{algorithm*}

\subsection{Learning Markovian and differentiable models from data}
\label{sec:methods:learning_markovian_and_differentiable_models_from_data}
The methods presented so far build on a key assumption: the availability of a high-quality Markovian system dynamics model that (i) is differentiable, and (ii) can be evaluated efficiently.

In cases where any of these assumptions is not met, we learn the required model\footnote{We write, e.g., \bobsseq\ for \bobsTs\ to reduce notational clutter.}
\eq{
	\p{\bobsseq}{\bactionseq} = \int &\p{\bobs_1}{\bstate_1} \, \initial{\bstate_1}\\& \prod_{t=2}^T \p{\bobst}{\bstate_t} \, \p{\bstate_t}{\bstate_{t-1} , \baction_{t-1}} \dint \bstateseq
}

from a data set of \emph{observations} and controls through a slight modification of the recently proposed Deep Variational Bayes Filters (DVBF; \cite{karls2016deep}); for details we refer to \cref{app:dvbf}.
This model has been shown to discover physically plausible system dynamics from raw sensor readings through the maximisation of the \emph{evidence lower bound} (ELBO) on the log-likelihood,
\eq{
    &\ln \p{\bobsseq}{\bactionseq} \geq\elbo(\bobsseq, \bactionseq)\ \\= 
        \, &\expc[q]{\ln \p{\bobsseq}{\bstateseq}} +\kl{\q{\bstateseq}{\bobsseq, \bactionseq}}{\p{\bstateseq}{\bactionseq}}.
}

By design, the learnt transition distribution can be reparametrised, as is necessary in \cref{sub:efficient_mi}.

When evaluating the learnt empowerment-maximising policy $\policy$ from \cref{sec:methods:empowerment_maximising_agents} in the real environment we make use of the observations and the recognition network $\q{\bstateseq}{\bobsseq, \bactionseq}$ from above.
This network acts as a filter for the latent state and keeps it from diverging from the true state of the environment.
This recognition network has the same complexity as the policy thus does not influence its real-time applicability.

\subsection{$n$-step empowerment}

Empowerment can also be computed over a horizon of actions and transition steps, it is then called \emph{$n$-step empowerment}.
Instead of maximising the mutual information between the taken action $\bactiont$ and the next state $\bstate_{t+1}$ given the current state $\bstatet$, we maximise the mutual information between the \emph{sequence} of $n$ actions $\baction_{t:t+n-1}$ and the consequential state $\bstate_{t+n}$, given the starting state $\bstatet$.
We can apply the same methodology as in the last section:
we simply switch from a single-step variational planning distribution $\q{\bactiont}{\bstatet, \bstate_{t+1}}$ to an $n$-step planning distribution $\q{\baction_{t:t+n-1}}{\bstatet, \bstate_{t+n}}$.

When experimenting $n$-step empowerment in section \cref{sec:experiments}, we keep the source distribution parametrised by one fully connected network, whereas we factorise the planning distribution, similarly to \cite{mohamed2015variational}:
\eq{
    &\q{\baction_{t:t+n-1}}{\bstate_{t+n}, \bstatet}\\ =\, &\q{\bactiont}{\bstate_{t+n}, \bstatet} \prod_{k=0}^{n-1} \q{\baction_{t+k+1}}{\bstate_{t+n}, \baction_{t+k}, \bstatet}.
}

%!TEX root = ../effemp.tex
\section{Related work}
\label{sec:related-work}
Since empowerment has been introduced \citep{klyubin2005empowerment,salge2014empowerment}, several works have attempted to provide a concrete and scalable implementation. 
The mutual information component is notoriously hard to calculate. 
Among other reasons, it requires integrating over all possible actions at all states, which is particularly difficult in continuous systems.
The early literature \citep{klyubin2005empowerment, klyubin2008options} circumvents these issues by operating in discrete-action state spaces and using the Blahut--Arimoto algorithm \citep{blahut1972computation} 
which extensively samples action and state space and uses a iterative scheme to update the source distribution. 
\cite{jung_empowerment_2011} take one step further by fitting the transitions with Gaussian processes and approximating empowerment by Monte-Carlo integration. 
They introduce empowerment for continuous state spaces but still needs discrete actions and furthermore requires a large sample size to provide satisfactory solutions. 
\cite{salge_approximation_2013} solve the intractability by linearising the dynamic systems which enables the use of well-known closed-form solutions.
More recently, \cite{mohamed2015variational} proposed a scalable approach by maximising a lower bound on the mutual information.
Their work does not assume or learn a model of the environment and is restricted to an open-loop setting. 
The term \emph{unsupervised control} was recently used by \cite{gregor2016variational}, who also resorted to an information-theoretic reward. 
Unlike \cite{mohamed2015variational}, the authors targeted closed-loop policies with two algorithms. 
The first, based on options \citep{sutton1999between}, is hard to train according to the authors and is restricted to linear function approximation. The second algorithm drops the option framework in order to overcome the aforementioned limitations. It is important to note that both algorithms are model-free, which constitutes the main difference with our approach.

Empowerment can be perceived as a member of a greater family of reward functions, reassembling intrinsic motivations \citep{singh2004intrinsically}.
We refer the reader to an excellent discussion on intrinsic motivation and a more thorough overview from \cite{oudeyer2008can};
here we present a few notable works.
Often the goal is to trigger explorative behaviour in the agent.
\cite{schmidhuber1991curious,schmidhuber2010formal} proposed to use a formalised notion of interestingness and curiosity.
\cite{bellemare2016unifying} show that increasing the value of under-visited states is related to information gain.
It has also been shown that a notion of \emph{surprise}, quantifying the degree to which an agent is observing something unexpected, can improve reinforcement learning \citep{itti2006bayesian,achiam2017surprise}.
One central notion in our work, the mutual information, has also been used to estimate information gain by \cite{houthooft2016vime} which is then integrated with the task-specific reward function.
Artificial creativity has been used to create games that aim to make an agent improve upon its current capabilities \citep{schmidhuber2013powerplay,sukhbaatar2017intrinsic}.
%!TEX root = ../effemp.tex
\section{Experiments}\label{sec:experiments}
We conduct four experiments to examine the method presented in \cref{sec:methods}.
The two first experiments (\cref{sub:bb,sub:pendulum}) verify the applicability of empowerment, even when (i) state and action are continuous, and (ii) the dynamics are unknown.
Moreover, we discuss the learnt behaviour when maximising empowerment.
\Cref{sub:mbb,sub:walker} examine two more complicated scenarios.
We determine whether the behaviour empowerment induce in simple environments consistently transfers to more interesting scenarios.

For reproducibility, the interested reader is referred to the appendix for details on the environments, system dynamics and computational architectures.

\subsection{Pendulum}
\label{sub:pendulum}
\begin{figure*}
	\centering
		\subfloat[Empowerment landscape and swing-up trajectory following the learnt policy. The colour represents the accumulated empowerment from \cref{eq:empowering_cost}.]{
			\includegraphics[width=.8\textwidth]{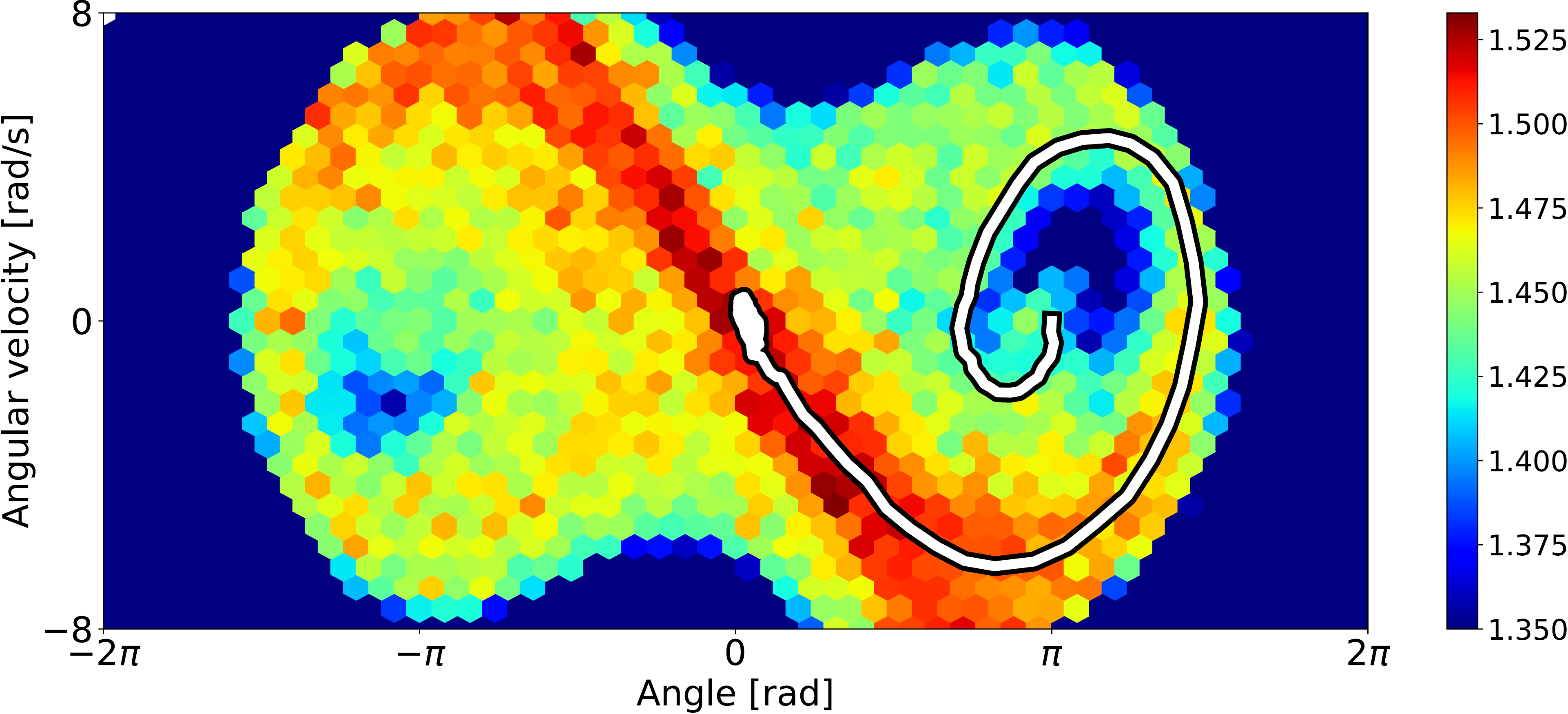}
			\label{fig:pendulum_empowerment}
		}\\
	\subfloat[Exemplary empowerment-based swingup sequence.]{
		\includegraphics[width=\textwidth]{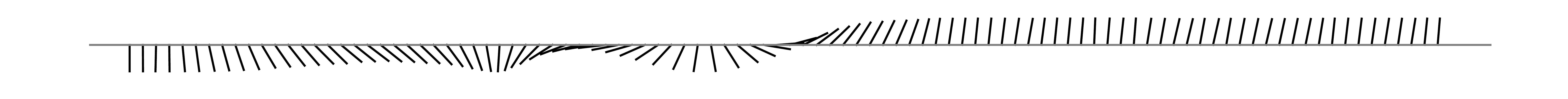}
		\label{fig:pendulum_sequence}
	}
	\caption{Results for the pendulum experiment.}
	\label{fig:pendulum}
\end{figure*}

Prior work investigated empowerment in a pendulum environment.  In particular, \cite{jung_empowerment_2011} discretise the action space;  \cite{salge_approximation_2013} linearise pendulum dynamics to make empowerment tractable.
We hence conducted an experiment in a continuous, torque-controlled pendulum environment as a proof of concept for our extended method. 
The observation space is two-dimensional---angle and angular velocity---and equal to the state space, with a one-dimensional torque control input.
The system dynamics are differentiable and explicitly known, cf.\ \cref{app:pendulum_dynamics}.

\Cref{fig:pendulum_empowerment} shows the empowerment landscape.
An intriguing property of this landscape is the cross-dependency between angle and angular velocity:
at the goal state, we want low velocity.
At intermediate steps, however, we need almost maximum velocity.
This would be particularly hard to model by hand---and impossible to generalise to other environments.

As the superimposed state-space walk of a swing-up from rest shows this simplifies finding optimal trajectories.
\Cref{fig:pendulum_sequence} visualises the same swing-up.
We observe that the learnt policy reliably swings up the pendulum.
This is in accordance with the aforementioned prior work on empowerment \citep{jung_empowerment_2011, salge_approximation_2013}:
in the upright position, we have most influence over the state.

\subsection{Single ball in a box}
\label{sub:bb}
\begin{figure*}
	\centering\captionsetup{width=.22\linewidth}
	\subfloat[Single ball in a box.]{
		\includegraphics[width=0.22\textwidth]{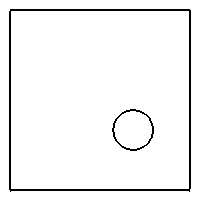}
	}
	\subfloat[Distribution of ball position when a random uniform policy is used. The balls tends to get stuck at walls and in corners.]{
		\includegraphics[width=0.22\textwidth]{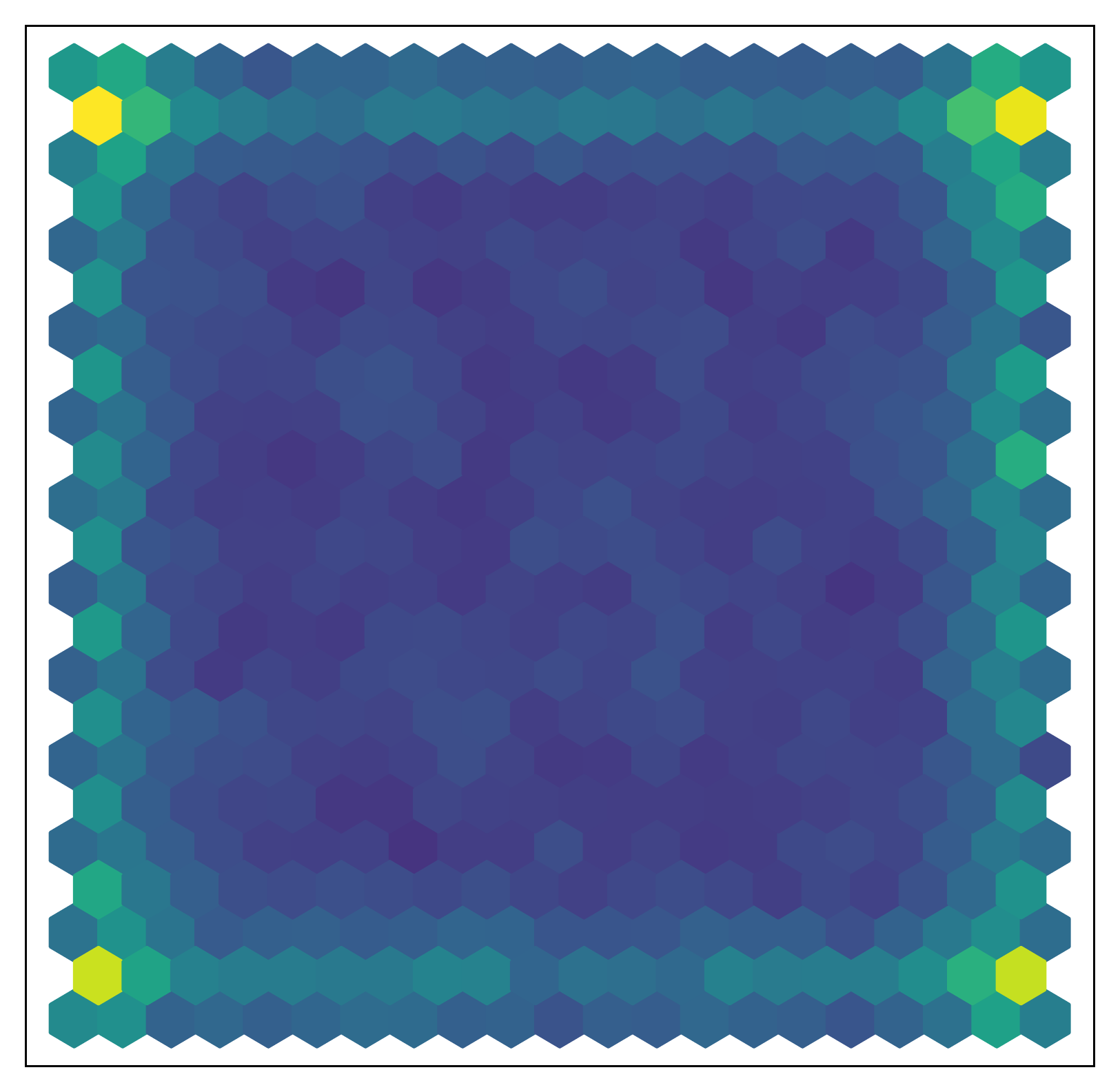}
		\label{fig:bb_random}
	}

	\subfloat[Distribution of ball position on em\-power\-ment-exploiting policy.]{
		\includegraphics[width=0.22\textwidth]{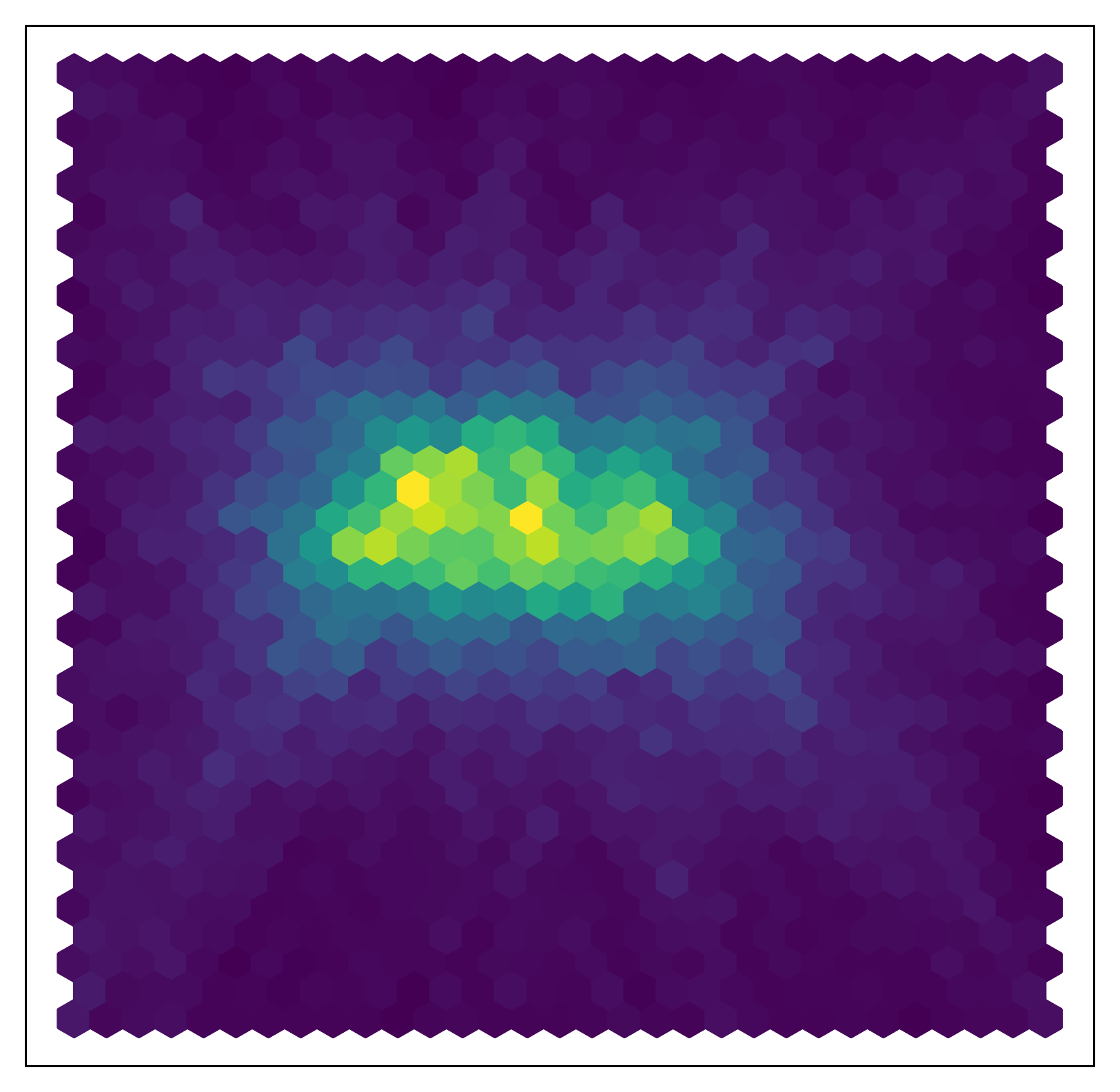}
		\label{fig:bb_empowered}
	}
	\subfloat[Distribution of ball position on 10-step em\-power\-ment-exploiting policy.]{
		\includegraphics[width=0.22\textwidth]{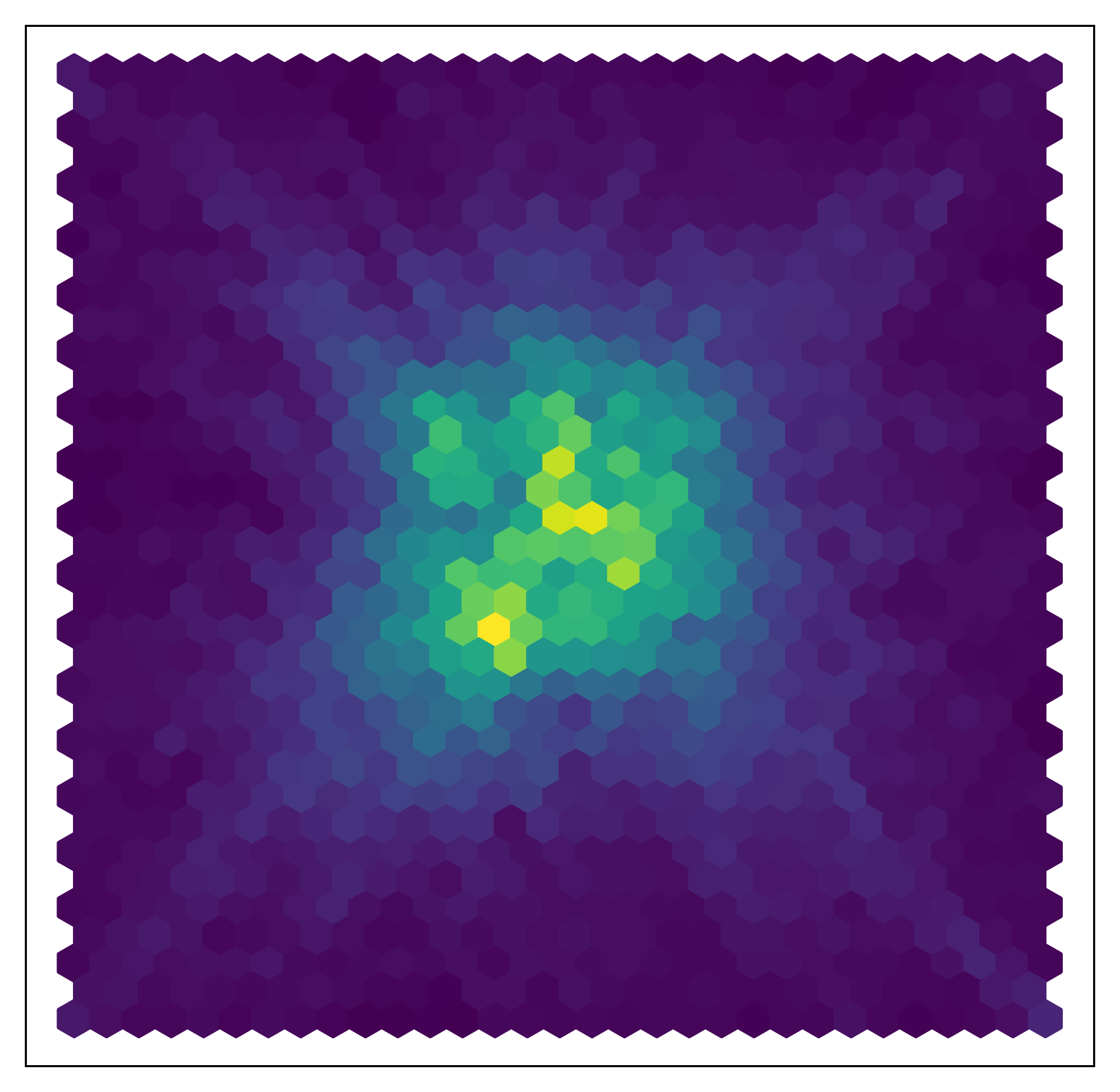}
		\label{fig:bb_empowered_nsteps}
	}
	\subfloat[Empowerment maximising policy. All actions move the ball towards the center of the box.]{
		\includegraphics[width=0.22\textwidth]{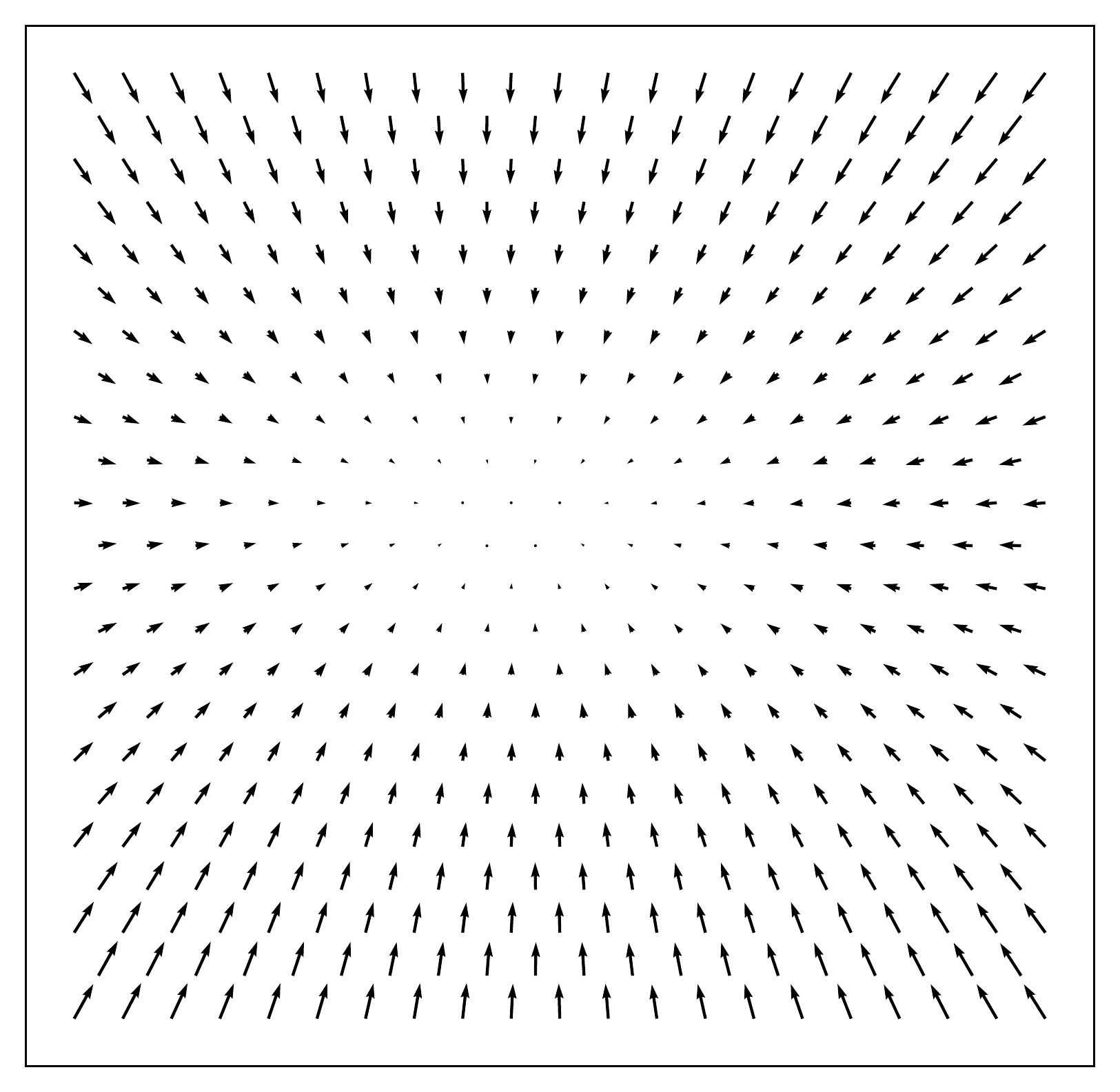}
		\label{fig:bb_policy}
	}
	\captionsetup{width=\linewidth}
	\caption{Results for the single ball in a box experiment.}
	\label{fig:bb_dist}
\end{figure*}
Next, we study a ball-in-a-box environment, which has been studied in previous literature \citep{wissner2013causal}.
The system dynamics are continuous but simple;
the two-dimensional controls are directly added to the two-dimensional position, which serves as the observation.
At a wall, any control perpendicular to that wall is absorbed.
In contrast to the pendulum experiment we do not assume the dynamics to be known and learn them according to \cref{sec:methods:learning_markovian_and_differentiable_models_from_data}.

We compare the behaviour of the empowerment-maximising policy $\policy$, depicted in \cref{fig:bb_policy} with a uniform policy in \cref{fig:bb_empowered,fig:bb_random}. 
We observe that the ball is controlled towards the centre of the bounding
box in the former case. 
In contrast, behaviour is suboptimal in the uniformly controlled case, as the ball tends to get stuck at the walls due to the absorption of control.

The policy plateaus in the centre of the box:
empowerment is constant within the region where the the ball cannot reach the wall within the horizon of \cref{eq:empowering_cost}---as long as it avoids getting stuck, the policy is unbiased.

\subsection{Multiple balls in a box}
\label{sub:mbb}
\begin{figure*}\centering\captionsetup{width=.22\linewidth}
		\subfloat[Environment snapshot with LIDAR visualisation for one ball.]{
			\includegraphics[width=0.22\textwidth]{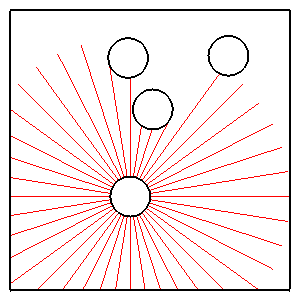}
			\label{fig:mbb}
		}%\hfill
		\subfloat[Policy averaged over all balls.]{
			\includegraphics[width=0.22\textwidth]{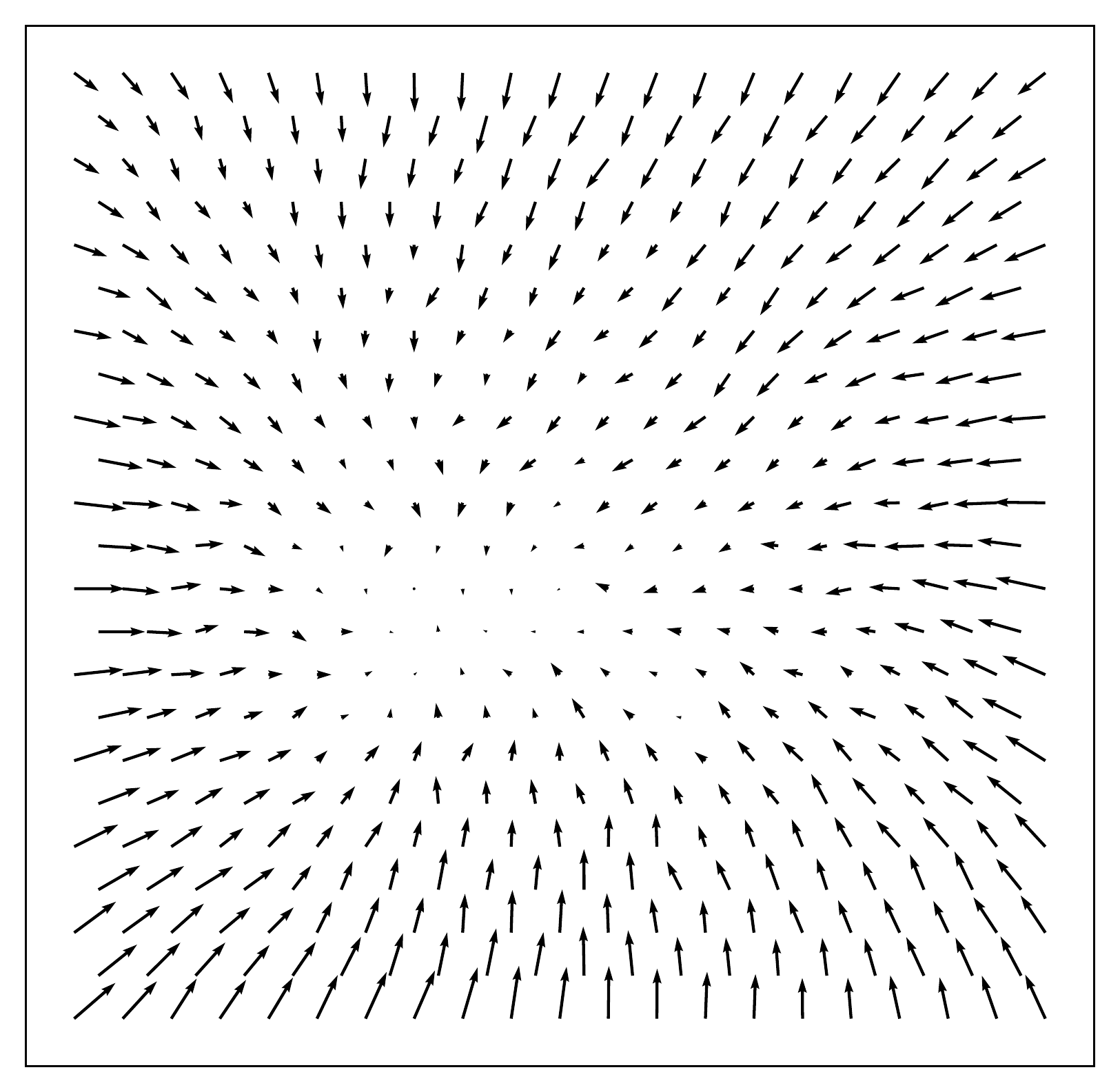}  
			\label{fig:mbb_policy}	
		}%\hfill
		\subfloat[Distance between ball and closest wall.]{
			\includegraphics[width=0.2219\textwidth]{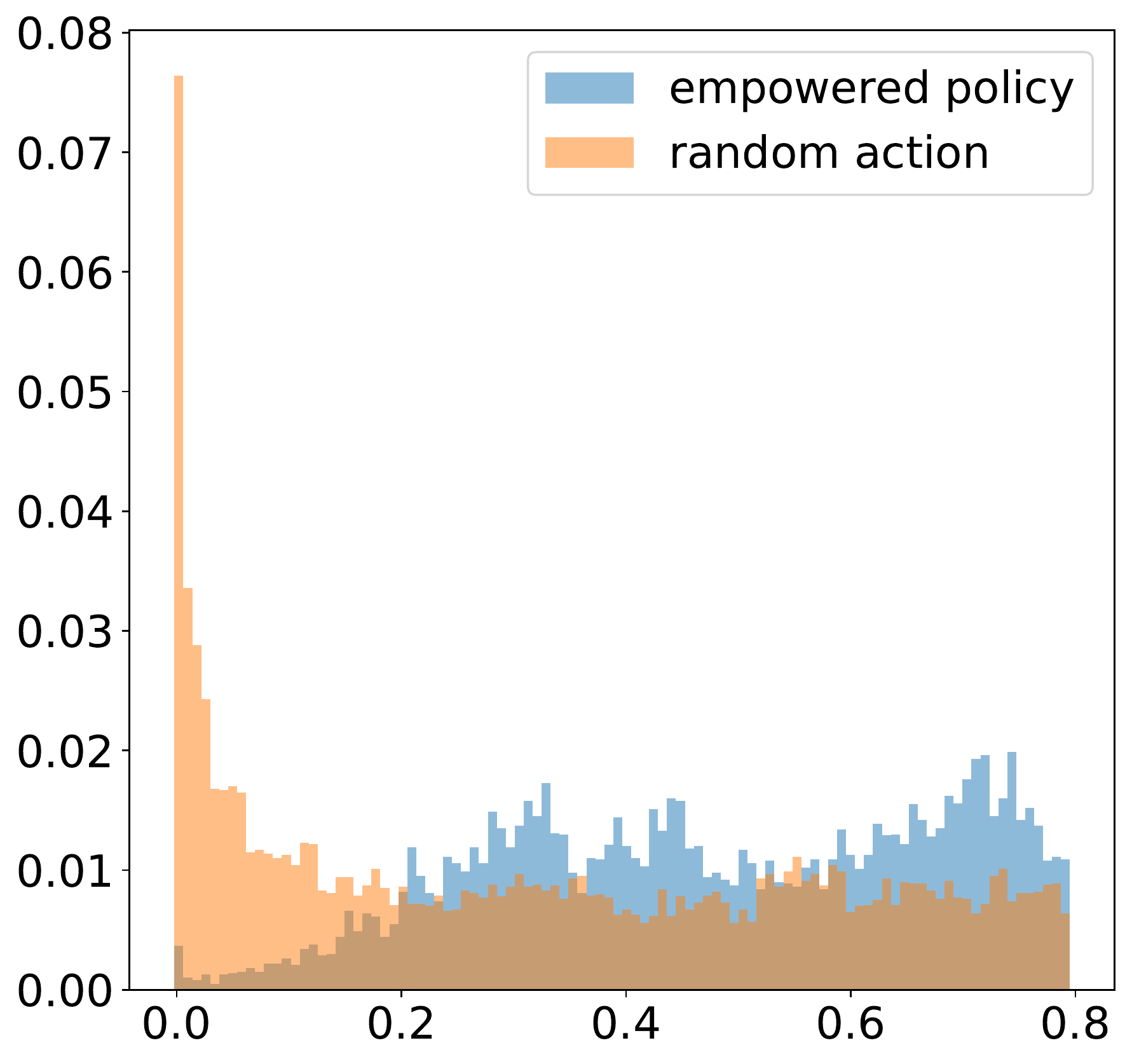}
			\label{fig:distancewall}
		}%\hfill
		\subfloat[Distance between two balls.]{
			\includegraphics[width=0.2277\textwidth]{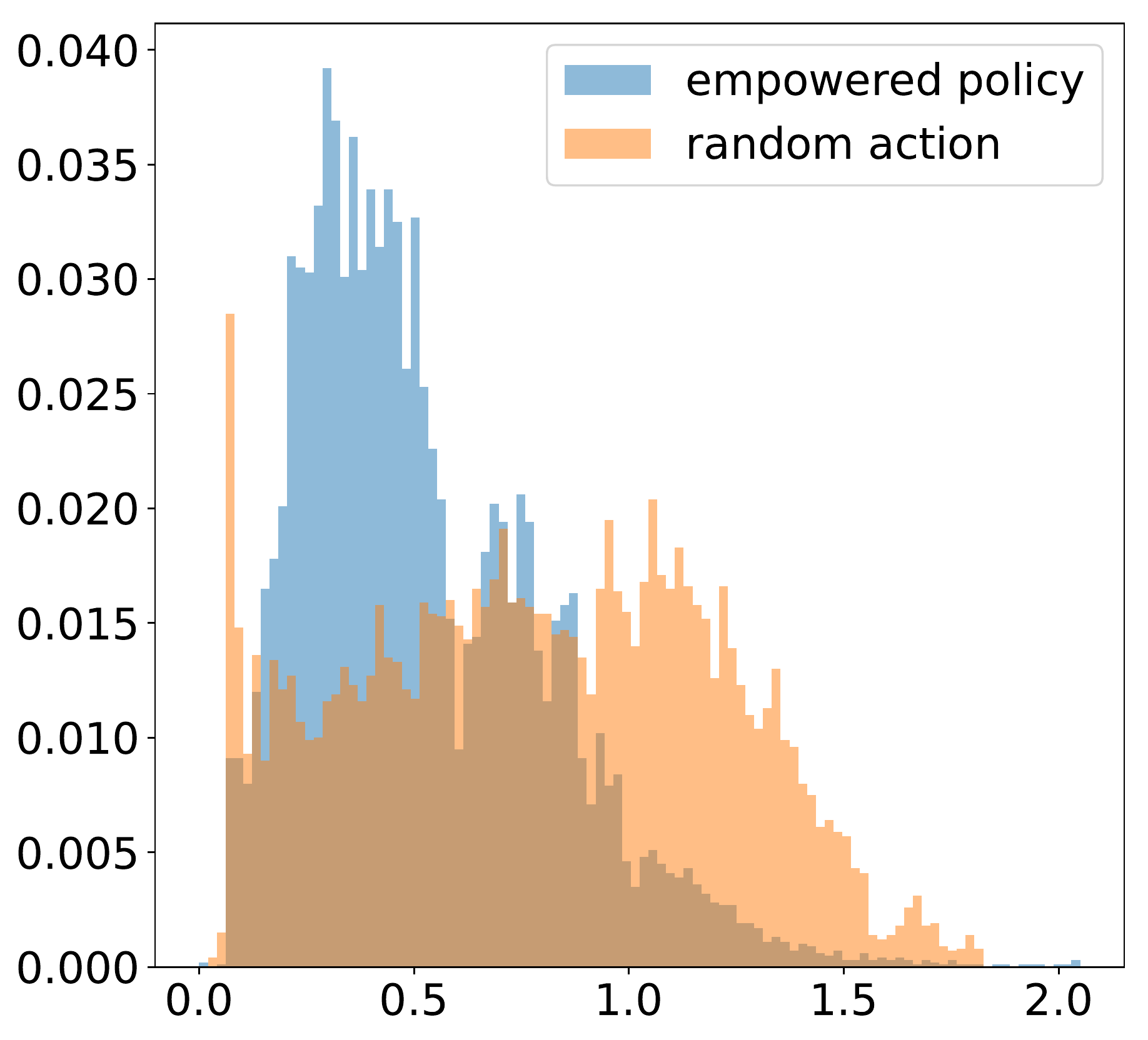}
			\label{fig:distanceballs}
		}
	\captionsetup{width=\linewidth}
	\caption{Results for the multiple balls in a box experiment.}
	\label{fig:mbb_hist}
\end{figure*}

In the multi-agent ball-in-a-box experiment, four balls interact in a bounding box.
Observation space and control of the balls are more complex compared to the previous environment.
Each has a \emph{continuous} 44-dimensional observation space: two position, two velocity, and 40 laser range (LIDAR) measurements.
Typical LIDAR distance signals for one ball are visualised in \cref{fig:mbb} as red lines.

A ball can be controlled with a two-dimensional force vector (proportional to acceleration), so the control is more intricate than in the previous environment with one ball.
Again, the walls are absorbing perpendicular forces.
Each ball has its own internal, 16-dimensional latent state-space model, as well as a  filter and a policy.
While the parameters of the transition model are shared and jointly trained, each ball computes and executes its action locally and independently.

The learnt policy $\policy$, \cref{fig:mbb_policy}, shows interesting properties.
Consider \cref{fig:distancewall}.
Compared to uniform sampling of actions, we observe that the balls try to avoid the walls.
This is expected from our previous experiment:
if a ball is close to a wall, all forces pointing towards the walls lead to negligible changes in its state:
empowerment decreases.

\Cref{fig:distanceballs} shows that this in turn leads to a decreased distance between balls.
On top, proximity by itself is beneficial to empowerment:
the changes in the LIDAR measurements are larger if the balls are closer to one another.
On closer observation of the histogram in \cref{fig:distanceballs}, we also find that they are less likely to bump into each other, for similar reasons as they avoid the walls.

The learnt policy hence consists in moving towards the centre of the box while avoiding collisions with the other balls.
Since the policy is shared, we can marginalise the effect of the other three balls by looking at the average action taken at a certain position.
The result is depicted in \cref{fig:mbb_policy}.
It shows the action taken at each position in the box, averaged over all balls.
This marginalises the effect of the relative position towards the other balls and allows us to compare this policy to the single-ball policy in \cref{fig:bb_policy}.
We see that we learn essentially the same policy---in an environment where the agent loses control over parts of the dynamics.

It should be stressed that the results given here are gathered after training.
This implies that empowerment is no longer evaluated.
It is amortised into a comparably small policy, cf.\ \cref{app:mbb_parameters}.

\subsection{Bipedal balancing}
\label{sub:walker}
\begin{figure*}
	\centering
	\includegraphics[width=1.0\textwidth]{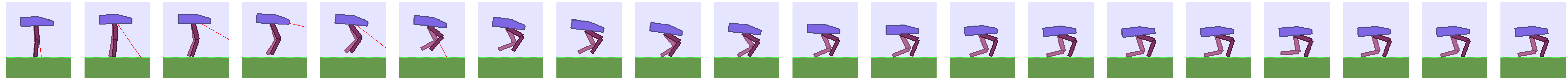}
	\caption{Bipedal walker balancing.}
	\label{fig:walker_balance}
\end{figure*}
In this environment, we control the two-joint legs of a biped robot.
Snapshots of the environment are visible in \cref{fig:walker_balance}.
The robot's leg joints are torque-controlled.

Observations include a total of 24 measurements provided by the environment: full state of the head, full state of each leg including ground contact, global velocity, and LIDAR rangefinder measurements. 
Again, we assume no prior knowledge on the dynamics; instead it is learnt with DVBF.

This environment is of particular interest due to inbuilt cost function in \cite{openaigym}:
the original goal is some sort of fast horizontal movement, a proxy to walking.
On top of rewarding forward movement, reward shaping is performed to punish (i) not keeping the head horizontal, (ii) falling, and (iii) excessive torque control.
These hard-coded and fine-tuned heuristic constraints tackle the problem that some form of balancing is a necessary prerequisite skill for walking.

Our experiments show that balancing can be unsupervisedly learnt via empowerment:
if the agent trips, it can henceforth only move its legs with no other effect---the influence on the environment is low, empowerment is decreased.

\subsection{$n$-step empowerment}
\begin{figure*}
	\centering
	\includegraphics[width=.75\textwidth]{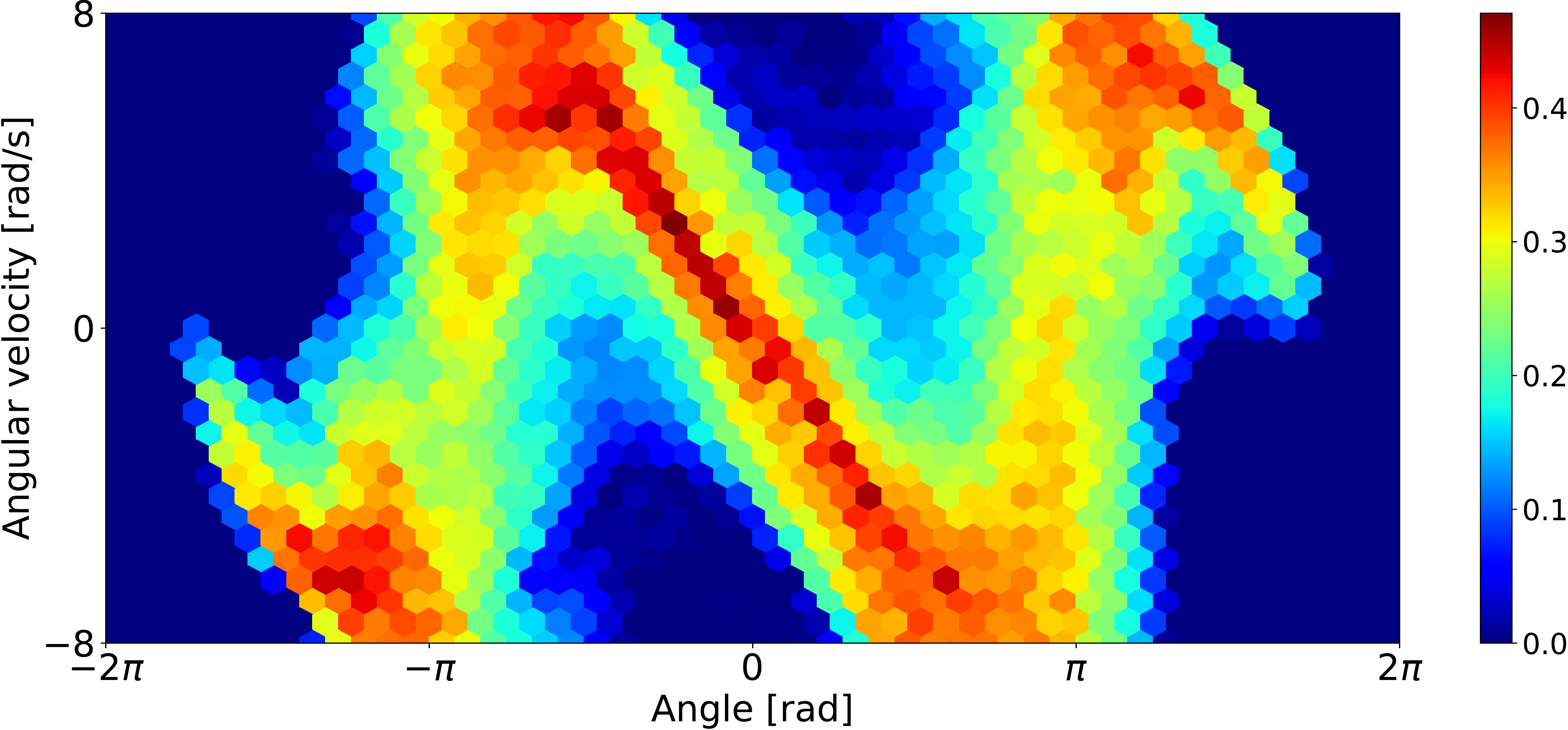}
	\label{fig:pendulum_empowerment_nsteps}
	\caption{Empowerment landscape for the 5-step empowerment function.}
	\label{fig:pendulum_nsteps}
\end{figure*}
We repeated two of the experiments with {$n$-step} empowerment to validate that the accumulated single-step rewards in \cref{eq:empowering_cost} can result in the same behaviour and the same cost landscape.
In the single ball experiment, we trained a policy to exploit 10-step empowerment and the accumulated single-step empowerment.
\Cref{fig:bb_empowered,fig:bb_empowered_nsteps} show the very similar distributions of the ball location when following the respective exploiting policies.

In the pendulum experiment, we used 5-step empowerment and compared it to the ordinary empowerment results from \cref{sub:pendulum}.
The 5-step empowerment landscape can be seen in \cref{fig:pendulum_nsteps}.
The results for both versions show a highly empowered state when the pendulum is standing upright and a path with monotone increasing empowerment towards this maximum.
\Cref{fig:pendulum_empowerment} shows a swing-up trajectory greedily following such a path.
The colors of \cref{fig:pendulum_empowerment} show accumulated one-step empowerment, \cref{eq:empowering_cost}, averaged over multiple rollouts, whereas the colors in \cref{fig:pendulum_nsteps} directly show $n$-step empowerment.
In both cases, the peak corresponds to a balanced pendulum.

\label{sub:pendulum_nsteps}

\section{Conclusion and future work}
\label{sec:conclusion_and_future_work}
Our experiments on unsupervised control in high-dimensional continuous dynamic systems demonstrate that empowerment leads to sensible basic behaviour, such as swinging up and balancing an inverted pendulum; distributing moving balls in a box; or balancing a biped.

The method is efficient in two ways.
First, its training procedure relies on stochastic gradient descent, which is known to gracefully scale  with the amount of model parameters and problem size.
Second, the result of the method---a policy maximising empowerment over a given time horizon---can be evaluated in constant time.
This makes the method real-time capable, a prerequisite for the deployment in many scenarios including robotics.

The flexibility of the model is inherited from its neural-network building blocks, where e.g.\ the statistical assumptions are encoded in the likelihood functions used.
Further, since training is done with gradient-based methods, the only requirement here is the differentiability of the loss functions.
We have proposed and shown that the major cause of inflexibility---the requirement for a differentiable, Markovian model of the system---can be  overcome by the estimation of a nonlinear Bayes filter from data only.

Future work includes the deployment of the method to physical systems as well as the integration of task-based reward functions, such as distance to a goal, with empowerment.

 \subsection*{Acknowledgments}
The authors would like to thank Jakob Breuninger for helpful discussions.
Our publication benefited from TensorFlow \citep{tensorflow}, and OpenAI Gym \citep{openaigym}.

\newpage
\bibliographystyle{apalike}
\bibliography{related}

\newpage
\onecolumn
\appendix
\section{Appendix}

\subsection{Tightness of the variational bound to mutual information}
\label{app:bound}
This section provides detailed steps omitted in \cref{eq:gap}:
\eq{
	\mi - \mibound 
	=& \iint \jointtrans \ln \frac{\planning}{\sourcetrans} \dint{\succstate} \dint{\nowaction} \\
	& \hspace{.5cm}- \iint \jointtrans \ln \frac{\apxplanning}{\sourcetrans} \dint{\succstate} \dint{\nowaction} \\
	=& \iint \jointtrans \ln \frac{\planning \cancel{\sourcetrans}}{\apxplanning \cancel{\sourcetrans}} \dint{\succstate} \dint{\nowaction} \\
	=& \int \margtrans \int \planning \ln \frac{\planning}{\apxplanning} \dint{\nowaction} \dint{\succstate} \\
	=& \  \expc[\succstate \sim \margtrans]{\kl{\planning}{\apxplanning}}.
	}

\subsection{Details on the experiments}
\subsubsection{General experiment settings}

Samples from policy or source distribution were transformed with the $\tanh$ function and then scaled to the environment's maximal possible action. We also used a KL divergence between the policy and a standard Normal distribution as a regulariser for the controls. When using the KL divergence the cost is scaled with the parameter $\beta$ to weight the influence of the regulariser.

\subsubsection{Pendulum}
\subsubsection*{Dynamics}
\label{app:pendulum_dynamics}
We used a TensorFlow implementation of the pendulum dynamics in OpenAI Gym \citep{openaigym} in order to be able to backpropagate through the transition. To be closer to prior work on empowerment and pendulum \citep{jung_empowerment_2011}, we added friction.
\eq{
        \Delta t &= 0.05\\
        g &= 10\\
        m &= 1\\
        l &= 1\\
        u &\leftarrow \min(\max(u, u_\mathrm{min}), u_\mathrm{max}) \\
        \dot\theta &\leftarrow \dot\theta + (-\frac{3  g }{ 2  l} \sin(\theta + \pi) + \frac{3 }{m  l^2}  (u - 0.05 \, \dot\theta)) \Delta t\\
        \theta &\leftarrow \theta + \dot\theta \Delta t\\
        \dot\theta &\leftarrow \min(\max(\dot\theta, -8), 8)
}

\subsubsection*{Parameters}
\label{app:pendulum_parameters}
Network parameters:
Number of action dimensions $n_u = 1$
\begin{itemize}
\item Policy $\policy{\baction_t}{\bstate_t}$: 128 tanh + 128 tanh + 128 tanh + 128 tanh + \{$n_u$ identity, $n_u$ exp\}
\item Source $\source{\baction_t}{\bstate_t}$: 128 tanh + 128 tanh + 128 tanh + 128 tanh + \{$n_u$ identity, $n_u$ exp\}
\item Planning $\policy{\baction_t}{\bstatet,\bstate_{t+1}}$: 128 tanh + 128 tanh + 128 tanh + 128 tanh + \{$n_u$ identity, $n_u$ exp\}
\item $\beta$ changing from 5 to 2000 over 800 epochs
\end{itemize}

\subsubsection{Single ball in a box}
\subsubsection*{Dynamics}
Single ball in a box simulated with Box2D based on OpenAI Gym environments.

Inelastic collisions, density 5.0, friction 1.1, radius 0.66
\label{app:bb_dynamics}

\subsubsection*{Parameters}
\label{app:bb_parameters}
Number of action dimensions $n_u = 2$\\
Number of latent dimensions $n_z = 32$
\begin{itemize}
\item Policy $\policy{\baction_t}{\bstate_t}$: 128 tanh + \{$n_u$ identity, $n_u$ exp\}
\item Source $\source{\baction_t}{\bstate_t}$: 128 tanh + \{$n_u$ identity, $n_u$ exp\}
\item Planning $\policy{\baction_t}{\bstatet,\bstate_{t+1}}$: 128 tanh + \{$n_u$ identity, $n_u$ exp\}
\item Recognition model $\recogmeas{\bstate_t}{\bobst}$: 128 relu + \{$n_z$ identity, $n_z$ square\}
\item Transition model $\transmean(\bstatet, \bactiont), \priortransvar(\bstatet, \bactiont), \inftransvar(\bstatet, \bactiont)$: 128 sigmoid + \{$n_z$ identity, $n_z$ square, $n_z$ square\}
\item Generative model $\p{\bobst}{\bstatet, \bactiont}$: \{128 relu + $n_x$ identity, 1 square\}
\item $\beta = 50$
\end{itemize}

\subsubsection{Multiple balls in a box}
\subsubsection*{Dynamics}
\label{app:mbb_dynamics}
Multiple balls in a box simulated with Box2D based on OpenAI Gym environments.

Inelastic collisions, density 5.0, friction 1.1, radius 0.66
\subsubsection*{Parameters}
\label{app:mbb_parameters}
Number of action dimensions $n_u = 2$\\
Number of latent dimensions $n_z = 16$
\begin{itemize}
\item Policy $\policy{\baction_t}{\bstate_t}$: 128 tanh + \{$n_u$ identity, $n_u$ exp\}
\item Source $\source{\baction_t}{\bstate_t}$: 128 tanh + \{$n_u$ identity, $n_u$ exp\}
\item Planning $\policy{\baction_t}{\bstatet,\bstate_{t+1}}$: 128 tanh + \{$n_u$ identity, $n_u$ exp\}
\item Recognition model $\recogmeas{\bstate_t}{\bobst}$: 128 relu + \{$n_z$ identity, $n_z$ square\}
\item Transition model $\transmean(\bstatet, \bactiont), \priortransvar(\bstatet, \bactiont), \inftransvar(\bstatet, \bactiont)$: 128 sigmoid + \{$n_z$ identity, $n_z$ square, $n_z$ square\}
\item Generative model $\p{\bobst}{\bstatet, \bactiont}$: \{128 relu + $n_x$ identity, 1 square\}
\item $\beta = 20$
\end{itemize}

\subsubsection{Bipedal robot}
\subsubsection*{Dynamics}
{\it BipedalWalker-v2} environment from OpenAI Gym.
\label{app:walker_dynamics}
\subsubsection*{Parameters}
\label{app:walker_parameters}
Number of action dimensions $n_u = 4$\\
Number of latent dimensions $n_z = 64$
\begin{itemize}
\item Policy $\policy{\baction_t}{\bstate_t}$: 128 tanh + \{$n_u$ identity, $n_u$ exp\}
\item Source $\source{\baction_t}{\bstate_t}$: 128 tanh + \{$n_u$ identity, $n_u$ exp\}
\item Planning $\policy{\baction_t}{\bstatet,\bstate_{t+1}}$: 128 tanh + \{$n_u$ identity, $n_u$ exp\}
\item Recognition model $\recogmeas{\bstate_t}{\bobst}$: 128 relu + \{$n_z$ identity, $n_z$ square\}
\item Transition model $\transmean(\bstatet, \bactiont), \priortransvar(\bstatet, \bactiont), \inftransvar(\bstatet, \bactiont)$: 128 sigmoid + \{$n_z$ identity, $n_z$ square, $n_z$ square\}
\item Generative model $\p{\bobst}{\bstatet, \bactiont}$: \{128 relu + $n_x$ identity, 1 square\}
\item $\beta = 60$
\end{itemize}

\subsection{Details on the models}
\subsubsection{Parametrisation of source and planning distributions}
\label{model:gaussmult}
Throughout this paper we chose $\source$ and $\q$ to be from restricted sets parameterised by $\sourcepars$ and $\planningpars$, respectively:
both are represented by Gaussian distributions of which the mean and the variance are parameterised by deep neural networks $\mu_{\sourcepars}, \sigma^2_{\sourcepars}, \mu_{\planningpars}$ and $\sigma^2_{\planningpars}$:
\eq{
	\source{\baction}{\bstate} =& \, \mathcal{N}(\mu_{\sourcepars}(\bstate), \sigma_{\sourcepars}^2(\bstate)), \\
	\apxplanning =& \, \mathcal{N}(\mu_{\planningpars}(\nowstate, \succstate), \sigma_{\planningpars}^2(\nowstate, \succstate)),
	}
	where $\sourcepars$ and $\planningpars$ are the respective weights.
	We denote the combination of $\sourcepars$ and $\planningpars$ as parameters $\sourceplanningpars$.

\subsubsection{Modifications to DVBF}
\label{app:dvbf}
We have used the Deep Variational Bayes Filter (DVBF; \cite{karls2016deep}) in many of the experiments described in Sec.~\ref{sec:experiments}.  Its deployment is explained here.

The initial state distribution $\initial{\bstate_1}$ is represented by a variational autoencoder, which we train along with the same objective. The latent state $\mathbf{w}_1$ of this initial autoencoder is created by a recognition model conditioned on the the first $K$ observations. This sample $\mathbf{w}_1$ is then transformed with another nonlinear function into the first latent state $\bstate_1$. The prior for this initial autoencoder is a standard Normal distribution.

The recognition model $\recog$ is implemented through two separate models which are applied at each time step $t$:
\eq{
	\recog{\bstate_t}{\bstate_{1:t-1}, \bobsts, \bactionts} \propto \recogmeas{\bstate_t}{\bobs_t} \times \recogtrans{\bstate_t}{\bstate_{t-1}, \baction_t},
}
where the belief over the current latent state given the last observation and the state history are implemented as the product of two distinct distributions, $\recogtrans$ and $\recogmeas$, respectively.
Both are implemented as Gaussian distributions.
As the product of their densities is a Gaussian density as well, arriving at $\recog$ is possible in closed form.
The exact implementation relies on parameter sharing with the prior:
\eq{
	\recogtrans{\bstate_{t}}{\bstate_{t-1}, \baction_{t-1}} =& \, \mathcal{N}\bigl(\transmean(\bstate_{t-1}, \baction_{t-1}), \inftransvar(\bstate_{t-1}, \baction_{t-1})\bigr), \\
	\recogmeas{\bstate_t}{\bobst} =& \, \mathcal{N}\bigl(\measuremean(\bobst), \measurevar(\bobst)\bigr).
}
Note that $\transmean$ is shared with the prior:
\eq{
    \p{\bstate_{t+1}}{\bstate_t, \baction_t} =
        \mathcal{N}\bigl(
                \transmean(\bstate_t, \baction_t),
                \priortransvar(\bstate_t, \baction_t)\bigr),
}
while $\inftransvar$ is not.
The resulting model then follows a Gaussian distribution \citep{murphy2012machine}: 
\eq{ 
	\q{\bstatet}{\bstate_{t-1}, \bobst, \baction_{t-1}} = \, \mathcal{N}(\recogmean, \recogvar),
}
with
\eq{ 
\recogmean = \frac{\transmean \measurevar + \measuremean \inftransvar}{\measurevar + \inftransvar}, \qquad
\recogvar = \frac{\measurevar \inftransvar}{\measurevar + \inftransvar}.
}

For easily integrating the change into the DVBF framework we compute the innovation noise variables as:
\eq{ 
\mathbf{\mu}_{\mathbf{w}_t} = \frac{\recogmean - \transmean}{\priortransstd}, \qquad
\mathbf{\sigma}^2_{\mathbf{w}_t} = \frac{\recogvar}{\priortransvar}
}

The KL divergence in the variational lower bound can then computed between $\mathcal{N}(\mathbf{\mu}_{\mathbf{w}_t}; \mathbf{\sigma}^2_{\mathbf{w}_t})$ and a standard Normal prior.

This allows the model to decide whether it should base its belief about $\bstate_t$ on $\bstate_{t-1}$ and $\baction_t$ or  on $\bobs_t$ instead.
All the mappings $\transmean, \priortransvar, \inftransvar, \measuremean, \measurevar$ are represented by feed-forward neural networks with a common parameter set $\modelpars$. 
The overall cost function for learning the model is 
\eq{
	-\expc[\bobsTs, \bactionTs \sim \empirical]{\elbo(\bobsTs, \bactionTs, \modelpars)} \approx
	-\frac{1}{N} \sum_{n=1}^N \elbo\left(\bobsTs^{(n)}, \bactionTs^{(n)}, \modelpars\right) =: \modelloss,
}
where $\empirical$ is the empirical distribution from which our data set $\dataset$ is obtained.
Estimation of the gradient is performed via backpropagation through time \citep{werbos1990backpropagation}, which lets us perform gradient descent on $\modelloss$ in $\modelpars$.

\end{document}